%% file: main.tex
\documentclass[10pt,twocolumn,letterpaper]{article}

\usepackage{cvpr}              

\input{preamble}

\definecolor{cvprblue}{rgb}{0.21,0.49,0.74}
\usepackage[pagebackref,breaklinks,colorlinks,citecolor=cvprblue]{hyperref}
\usepackage[accsupp]{axessibility}
\def\paperID{3278} 
\def\confName{CVPR}
\def\confYear{2024}

\title{NTO3D: Neural Target Object 3D Reconstruction with Segment Anything}

\author{
    {\normalsize Xiaobao Wei$^{1,2,3}$ \quad Renrui Zhang$^{4}$ \quad Jiarui Wu$^{1,3}$ \quad Jiaming Liu$^{1}$}\\ 
    {\normalsize Ming Lu$^{5}$ \quad Yandong Guo$^{6}$ \quad Shanghang Zhang$^{1\dagger}$}\\
    {\normalsize{$^{1}$National Key Laboratory for Multimedia Information Processing, School of Computer Science,}}\\
    {\normalsize Peking University \quad $^{2}$Institute of Software, Chinese Academy of Sciences}\\
    {\normalsize $^{3}$University of Chinese Academy of Sciences \quad $^{4}$Shanghai Artificial Intelligence Laboratory}\\
    {\normalsize $^{5}$Intel Labs China \quad $^{6}$AI$^2$ Robotics}\\
}

\begin{document}
\maketitle
\begin{abstract}
Neural 3D reconstruction from multi-view images has recently attracted increasing attention from the community. Existing methods normally learn a neural field for the whole scene, while it is still under-explored how to reconstruct a target object indicated by users. Considering the Segment Anything Model (SAM) has shown effectiveness in segmenting any 2D images, in this paper, we propose NTO3D, a novel high-quality Neural Target Object 3D (NTO3D) reconstruction method, which leverages the benefits of both neural field and SAM. We first propose a novel strategy to lift the multi-view 2D segmentation masks of SAM into a unified 3D occupancy field. The 3D occupancy field is then projected into 2D space and generates the new prompts for SAM. This process is iterative until convergence to separate the target object from the scene. After this, we then lift the 2D features of the SAM encoder into a 3D feature field in order to improve the reconstruction quality of the target object. NTO3D lifts the 2D masks and features of SAM into the 3D neural field for high-quality neural target object 3D reconstruction. We conduct detailed experiments on several benchmark datasets to demonstrate the advantages of our method. The code will be available at: \href{https://github.com/ucwxb/NTO3D}{https://github.com/ucwxb/NTO3D}.
\end{abstract}
\renewcommand{\thefootnote}{\fnsymbol{footnote}} 
\footnotetext[2]{Corresponding Author E-mail: shanghang@pku.edu.cn}

\section{Introduction}

The neural field has made significant progress over the past few years and become one of the most popular 3D representations. The pioneering Neural Radiance Field (NeRF)~\cite{mildenhall2021nerf} and its variants~\cite{muller2022instant,sun2022direct,barron2021mip,yu2021plenoctrees,fridovich2022plenoxels} learn coordinate-based neural networks to predict the density and color from multi-view images and use volume rendering to conduct novel view synthesis. NeuS~\cite{wang2021neus} improves the 3D reconstruction quality of NeRF by representing a surface with a Signed Distance Function (SDF). They also developed a new volume rendering method to train the neural SDF representation. Many studies are proposed to improve the reconstruction quality and reduce the training cost~\cite{wu2022voxurf,tang2023delicate}. However, existing methods usually learn a neural field for the whole scene, ignoring the reconstruction of a target object in the scene, which can be indicated by end users on the fly. 
\begin{figure}
    \centering
    \includegraphics[width=1.0\linewidth]{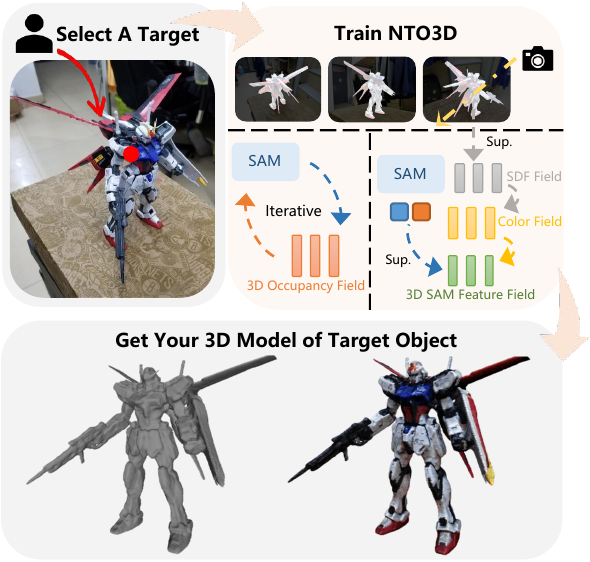}
    \caption{\textbf{Overview of NTO3D.} First, a user selects a reconstruction target in the scene. Then, our NTO3D utilizes a 3D occupancy field iteratively to merge the multi-view 2D segmentation masks into 3D space. NTO3D further lifts the features of the SAM encoder into a 3D SAM features field and optimizes the feature field together with other fields. Finally, the user can obtain a high-quality 3D reconstruction model of the target object with NTO3D.}
    \label{fig:intro}
\end{figure}

Although traditional techniques such as in-hand scanning~\cite{weise2009hand} have been proposed for target 3D object reconstruction, it is still non-trivial for neural 3D reconstruction methods since we need to obtain the multi-view consistent target object segmentation, which is labor-intensive and time-consuming.    

Recently, the Segment Anything Model (SAM)~\cite{kirillov2023segment} has shown great potential for zero-shot segmentation, which can be used to segment a target object out of the scene. However, with a single prompt, SAM can only obtain the 2D segmentation of a single-view image, other than multi-view images. 
In addition, how to leverage the features of SAM to improve the reconstruction quality is still under-explored to the best of our knowledge.



To address the above issue, we propose NTO3D, a novel high-quality Neural Target Object 3D (NTO3D) reconstruction method that fully leverages the benefits of both neural field and SAM. Specifically, to separate the target object from the neural field, we first train a 3D occupancy field to merge the multi-view 2D segmentation masks. Our 3D occupancy field is based on the following assumptions: (1) If a pixel is foreground, then at least one of the positions passed through the ray is foreground. (2) If a pixel is background, then all the positions passed through the ray are background. We design a corresponding loss based on the assumptions to optimize the 3D occupancy field, lifting the 2D masks to a unified 3D occupancy field. The 3D occupancy field is then projected into 2D space and generates the new prompts for SAM. This process is iterative until convergence to finally segment the target object out of the scene. 

After this, in order to improve the reconstruction quality of the target object, we further lift the features of the SAM encoder into a 3D feature field. We add a lightweight output head to the neural field for learning the 3D features of SAM and use volume rendering to render the 2D features. The rendered 2D features are directly supervised by the 2D features of SAM. By lifting the 2D features of SAM, our method can reconstruct a more accurate 3D model for the target object.

Our main contributions are summarized as follows:

\begin{itemize}
\item We propose NTO3D, a novel method that iteratively lifts the 2D masks of SAM into a unified 3D occupancy field, segmenting the target object out of the neural field. With our method, users can easily reconstruct any target objects by prompting in a single view.

\item To boost the reconstruction quality, we further present a tactful strategy to lift the 2D features of SAM into a 3D feature field.

\item We conduct detailed experiments on DTU, LLFF, and BlendedMVS datasets, where NTO3D surpasses the state-of-the-art reconstruction methods, demonstrating the advantages of our approach.
\end{itemize}

\section{Related Work}


{\bf Neural Implicit Representation.} Neural implicit representation has recently become prevailing in computer vision and graphics. This representation utilizes coordinate-based neural networks to represent a field, which can encode continuous signals of arbitrary dimensions at arbitrary resolutions. Neural implicit representation has shown promising results in shape reconstruction~\cite{mescheder2019occupancy,Michalkiewicz_2019_ICCV,park2019deepsdf,8953765,atzmon2020sal,gropp2020implicit,yifan2020iso,Peng2020ConvolutionalON}, novel view synthesis~\cite{sitzmann2019scene,lombardi2019neural,kaza2019differentiable,mildenhall2020nerf,liu2020neural,saito2019pifu,saito2020pifuhd, trevithick2020grf, sitzmann2019deepvoxels} and multi-view 3D reconstruction~\cite{yariv2020multiview,niemeyer2020differentiable,kellnhofer2021neural,jiang2020sdfdiff, liu2020dist}. In particular, Neural Radiance Fields (NeRF)~\cite{mildenhall2021nerf} learn a continuous volume density and radiance field from multi-view images. After training, it can render images from arbitrary views via volume rendering. To improve the surface reconstruction quality of NeRF, NeuS~\cite{wang2021neus} utilizes a Signed Distance Function (SDF) to represent a surface. Voxurf and NeRF2Mesh\cite{wu2022voxurf,tang2023delicate} are proposed to reduce the training cost and improve the reconstruction quality. However, how to effectively reconstruct a target object with neural implicit representation is still under-explored, as it is difficult to obtain multi-view consistent target object segmentation. Though SA3D~\cite{cen2023segment} achieve neural rendering of target objects with SAM, it fails to impose geometry constraints for neural reconstruction. 
In this paper, we proposed a unified 3D occupancy field to effectively segment a target object out of the neural field.

{\bf Image Segmentation.} Great efforts have been made for different segmentation tasks such as semantic segmentation~\cite{cheng2022masked}, instance segmentation~\cite{ tian2020conditional}, and panoptic segmentation~\cite{kirillov2019panoptic}. Various models have also been developed for segmentation, including encoder-decoder structures~\cite{ronneberger2015u}, dilated convolutions~\cite{yu2015multi}, pyramid structures~\cite{zhao2017pyramid} and transformers~\cite{xie2021segformer}.
Recently, the Segment Anything Model (SAM)~\cite{kirillov2023segment} and its variants~\cite{ wang2023seggpt, zhang2023personalize} have demonstrated strong zero-shot generalization ability, enabling 2D segmentation for diverse real-world target objects. However, SAM is currently limited to 2D segmentation of a single image, which is insufficient for multi-view consistent target object segmentation. Moreover, utilizing features of SAM for improving the 3D reconstruction quality remains under-explored. In this study, we lift the 2D masks and features of SAM into the 3D neural field for high-quality neural target object 3D reconstruction.

{\bf 3D Reconstruction.} The problem of 3D reconstruction has been extensively studied in computer vision with numerous methods proposed for various applications. Traditional RGB-based methods usually rely on multi-view stereo techniques to predict the depth from posed images~\cite{furukawa2015multi,seitz2006comparison}. Recent learning-based methods aggregate multi-view information to learn the 3D representation of the scene~\cite{yao2018mvsnet,yao2019recurrent}. With the development of neural implicit representation, current methods start to represent the 3D scene with various neural fields~\cite{wang2021neus,mildenhall2020nerf}. Plenty of methods are also proposed for reconstructing various specific objects such as 3D faces~\cite{beeler2010high,beeler2011high}, bodies~\cite{zhu2016video}, and hands~\cite{wang2013video}. Apart from RGB cameras, many methods are proposed to use more cameras for 3D reconstruction. For example, KinectFusion~\cite{izadi2011kinectfusion} enables a user to rapidly create detailed 3D reconstruction by holding and moving a standard RGB-D camera. VoxelHashing~\cite{niessner2013real} improves the regular grid data structure of KinectFusion with a simple spatial hashing scheme that compresses space. AutoRecon~\cite{wang2023autorecon} leverages self-supervised 2D vision transformer features and reconstruct decomposed neural scene representations with decomposed point clouds, to achieve accurate object reconstruction and segmentation. Although 3D reconstruction is a well-studied problem, how to reconstruct a certain object indicated by users on the fly is still a difficult problem. By leveraging the benefits of both neural fields and SAM, our method enables users to easily reconstruct any target objects by prompting in a single view.    

\begin{figure*}[!ht]
    \centering
    \includegraphics[width=0.9\textwidth]{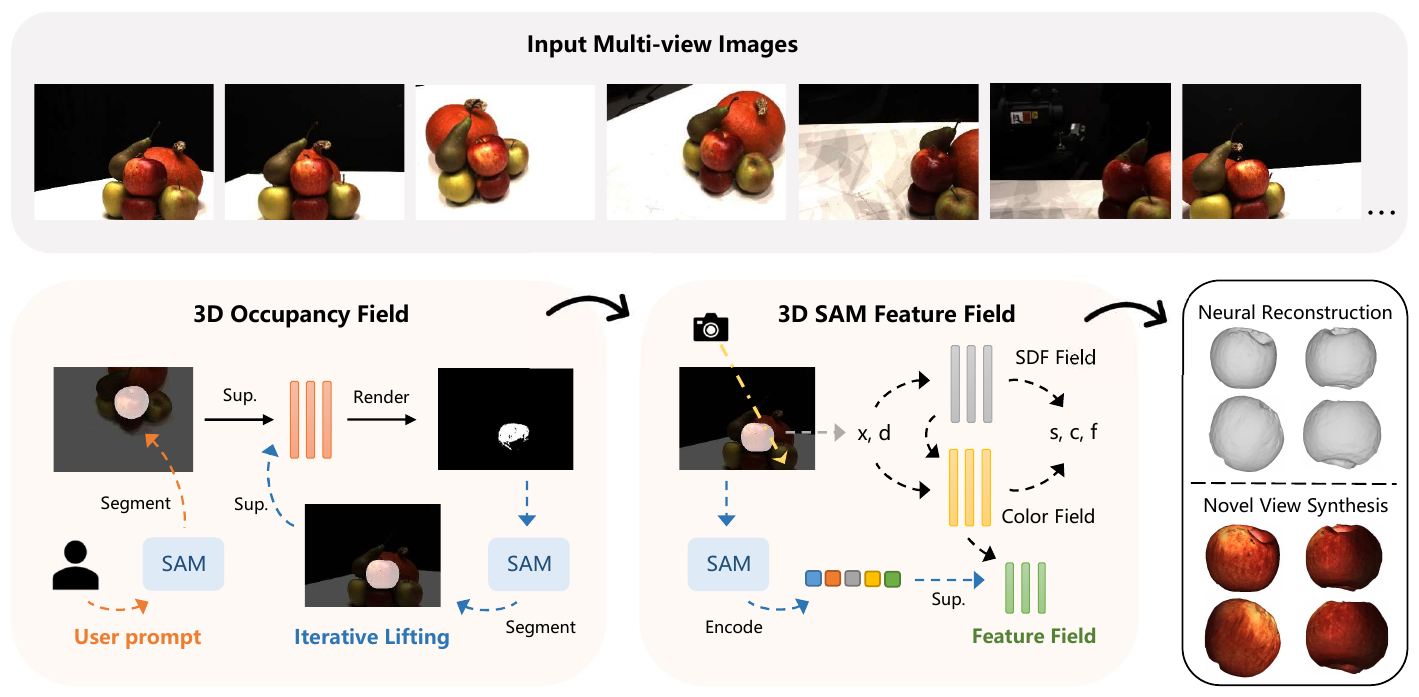}
    \caption{The overall pipeline of NTO3D. First, the user specifies the target object to be reconstructed and sends prompts to SAM for segmentation on the initial view. With multi-view images as input, we train the 3D occupancy field iteratively to lift cross-view masks into 3D space. When the 3D occupancy field converges to high-quality masks of the target objects, we finetune the pre-trained neural field based on the masked images and distill SAM encoder features into 3D space to obtain better reconstruction quality.}
    \label{fig:pipeline}
\end{figure*}

\section{Method}
In this section, we first briefly review neural object 3D reconstruction. Subsequently, we proceed to elaborate on the pipeline of the proposed Neural Target Object 3D  (NTO3D). Finally, we further elucidate the novel designs incorporated in NTO3D.
\subsection{Preliminaries}
Recent neural object 3D reconstruction works such as NeRF~\cite{mildenhall2021nerf} and NeuS~\cite{wang2021neus} both learn coordinate-based neural networks to represent the scene. NeRF constructs a mapping function from spatial location $x \in \mathbb{R}^3$ and view direction $d \in \mathbb{R}^2$ to color $c\in\mathbb{R}^3$ and volume density $\sigma$. Different from NeRF, NeuS replaces the density field with a signed distance field. We can extract the geometry surface $\mathbb{S}$ of the scene by the zero-set of the SDF values $S= \{ x \in\mathbb{R}^3| f_{sdf}(x)=0 \}$, where $ f_{sdf} $ is the signed distance function.
Based on the signed distance function, we can further calculate the opaque density $\rho$ and opacity values $\alpha$. Finally, the pixel color $\hat{C}$ of a ray $t$ can be computed by the classical volume rendering function:
\begin{equation}
\label{eq:volume_render}
    \hat{C}(t) =\sum_{i=1}^n T(t_i)\alpha(t_i)\mathbb{c}(t_i)
\end{equation}
where $n$ is the number of sample points along one ray and $T$ represents the discrete accumulated transmittances, which is defined as $T_i=\Pi_{j=1}^{i-1}(1-\alpha_j)$. 

\subsection{Overall Pipeline}
As shown in Fig.~\ref{fig:pipeline}, our method consists of two stages. In the first stage, we train the neural field of the scene on multi-view images. The users can then indicate the prompts of a target object in a single view. We use SAM to obtain the segmentation of this view and initialize the 3D occupancy field. The 3D occupancy field can generate coarse masks for other views and further aggregate to the prompts of SAM. Precise masks of other views can be generated by SAM to refine the initialized 3D occupancy field. This process is iterative until the 3D occupancy field converges. 

In the second stage, after segmenting the target object out of the scene, we can obtain the precise 2D masks of the target object in all views. We then leverage the features of SAM to improve the reconstruction quality of the target object, by distilling the 2D features of the SAM encoder into the 3D feature field. We will present the two stages in the following sections. 
\subsection{Stage-1: Segmentation by 3D Occupancy Field}
\begin{figure}[ht]
    \centering
    \includegraphics[width=0.8\linewidth]{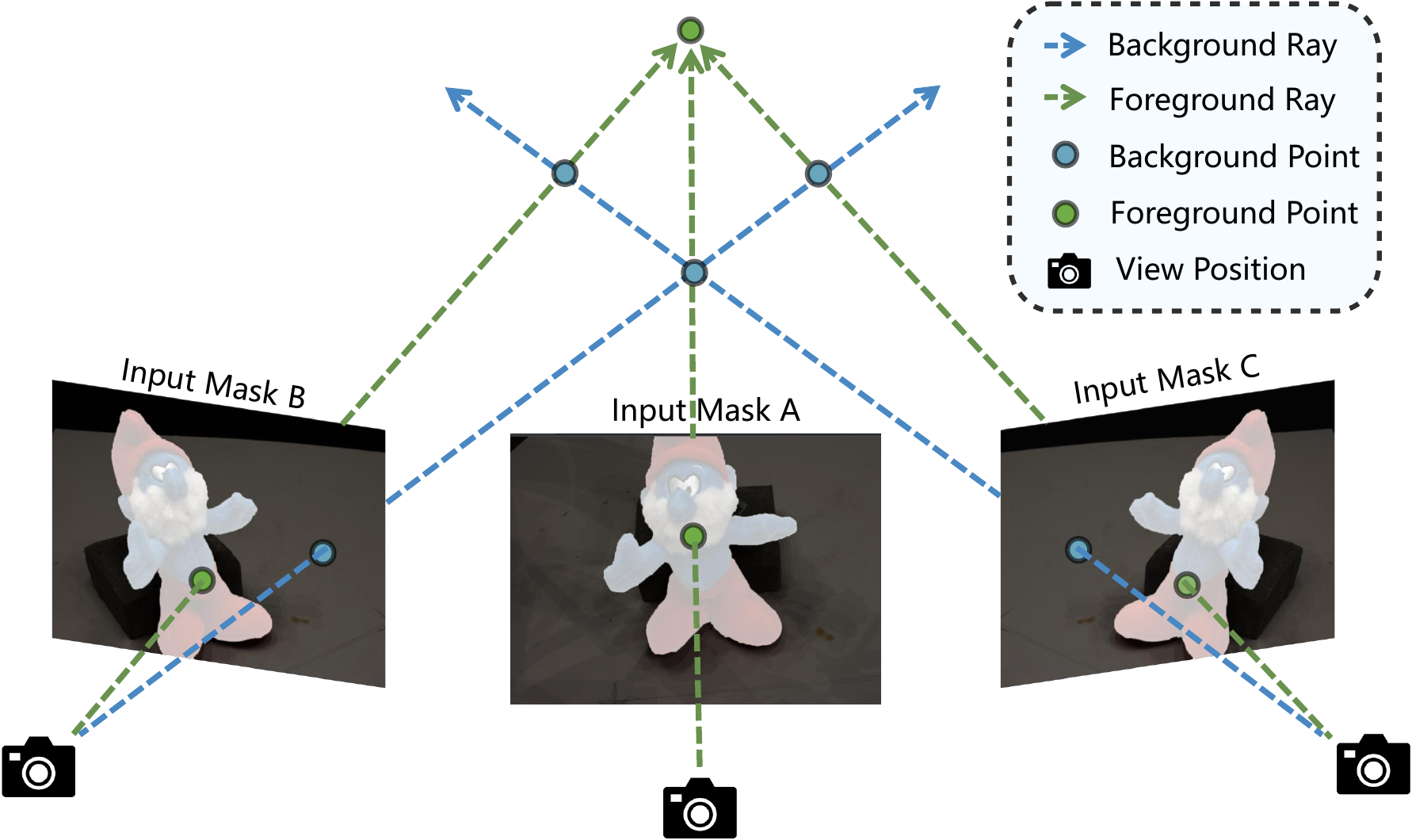}
    \caption{The illustration of the 3D occupancy field. Implicit interaction between multiple rays to decide which point is foreground or background. For a background ray, all points on it belong to the background. For a foreground ray, at least one point on it is foreground. }
    \label{fig:NOC_unified_field}
\end{figure}

In this section, we introduce a 3D occupancy field to lift 2D segmentation masks from different views into 3D space as shown in Fig.~\ref{fig:NOC_unified_field}. The 3D occupancy field can be used to identify foreground and background voxels and generate a unified 3D segmentation mask of the target object. By constructing the 3D occupancy field, we can also obtain multi-view consistent 2D masks within a short time.

Given a coarse 2D mask rendered by the 3D occupancy field, we use SAM to refine it. Although SAM supports a variety of prompts as input, including points, boxes, masks, and text, we use points and boxes as prompts, which works better and saves computational memory. We compute the K-means clustering centers and minimum bounding rectangles of a 2D coarse mask as prompts. Then SAM encodes image $I$ and prompts $P=(Point,Box)$ as features and decodes features into more precise masks $M_{SAM}$. The process can be formulated as:
\begin{equation}
\label{eq:mask_to_sam}
    M_{SAM}=Dec_{M}(Enc_{I}(I),Enc_{P}(P))
\end{equation}
where $Enc_{I}$, $Enc_{P}$, and $Dec_{M}$ are the image encoder, prompts encoder, and mask decoder in SAM, respectively. $M_{SAM}$ will be used as the label to supervise the learning of the 3D occupancy field.

Given a ray $r(t)=o+t \cdot d$ with camera position $o$ and view direction $d$, the corresponding 2D segmentation is defined as $M(r)$. In the box of the scene, the majority of voxels belong to the background, with only a small portion belonging to the foreground. Therefore, the relations between the pixel-level segmentation $M$ and the 3D occupancy field $M_{o}$ are based on the following two assumptions: (1) If a pixel belongs to the foreground, then at least one of the positions passed through the ray is foreground. (2) If a pixel belongs to the background, then all the positions passed through the ray are background. 
With the above assumptions, we can simply apply maximization operation to formulate the relations between $n$ points along the ray $r$:
\begin{equation}
\label{eq:max_voxel}
M(r)=f_{s}(\max(\{M_{o}(r(t_i)) \cdot \omega_{r(t_i)}|i\in\{1,\dots ,n\}))
\end{equation}
where $f_{s}$ indicates the sigmoid function and $\omega_{r(t_i)}$ represents point-wise weights, which are in stop-gradients from the SDF field. Then we can train the 3D occupancy field by the binary cross-entropy loss:
\begin{equation}
\label{eq:variation_loss}
    L_{o}=L_{BCE}(M_{SAM}, M)
\end{equation}

\begin{figure}[t]
    \centering
    \includegraphics[width=1.0\linewidth]{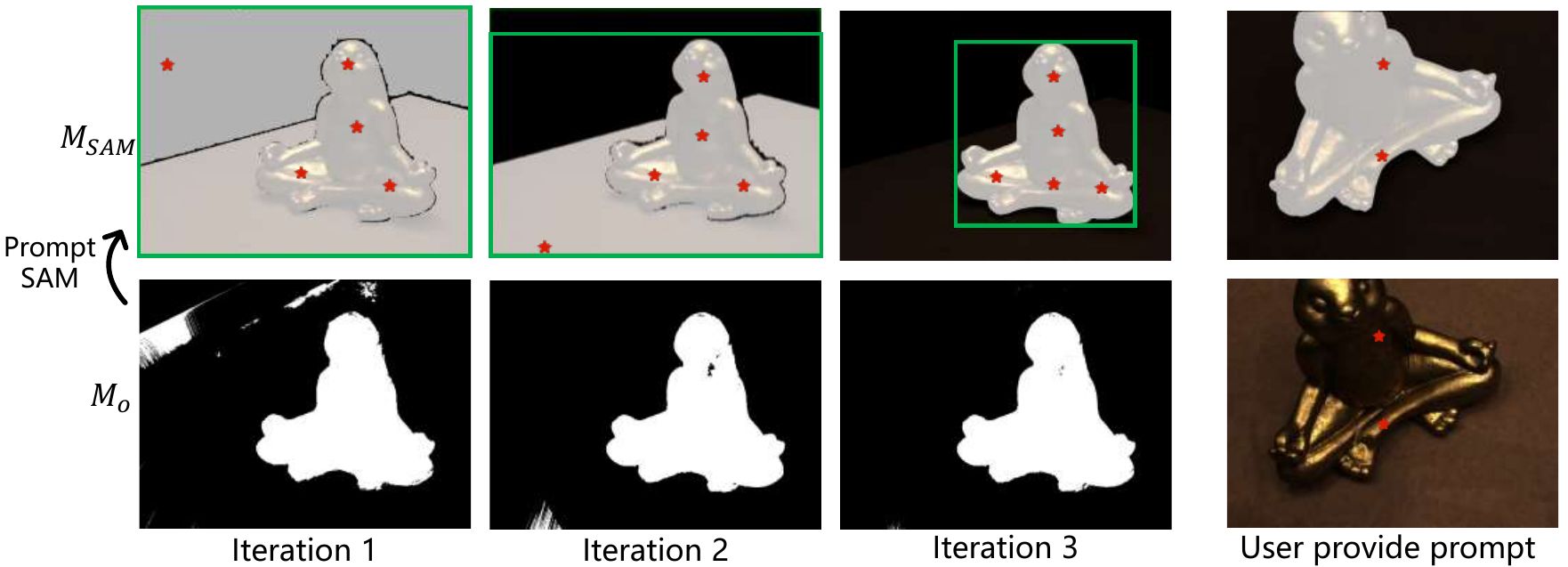}
    \caption{Mask iteratively lifting illumination, in which $M_{o}$ represents masks generated by 3D occupancy field and $M_{SAM}$ indicates masks provided by SAM base on prompts. Given users' prompts of specific objects, the 3D occupancy field renders a coarse mask in another view, which leads to bad prompts for SAM and defective masks. But the 3D occupancy field lifts 2D masks from all views into 3D space and efficiently corrects its false judgments of voxels in other views. With the iterative training, $M_{o}$ and $M_{SAM}$ begin to shrink and finally converge to the same.}
    \label{fig:Iteration_mask}
\end{figure}

By minimizing the above loss function, we can transfer the foreground segmentation of SAM to a unified 3D occupancy field. The above process is iterative until the 3D occupancy field converges. As shown in Fig.~\ref{fig:Iteration_mask}, at the beginning of the iteration, the rendered 2D coarse masks may exhibit defects since the 3D occupancy field does not converge. However, the refined 2D precise masks are mostly correct due to the proper prompting of SAM. After several iterations, the 3D occupancy field can correct erroneous predictions, ensuring multi-view consistent 2D masks.
\subsection{Stage-2: Refinement by 3D Feature Field}

As a foundation segmentation model, SAM possesses the ability to surpass most previous segmentation models and contains abundant knowledge. To better leverage the features of SAM, we add a lightweight output branch to the neural field, lifting the features of the SAM encoder into 3D space. With the encoder of SAM, we can obtain feature maps $Enc_{I}(I)$ with $256 \times 256$ resolutions corresponding to the input images. Then this branch takes the 3D point position, geometry features, and color features as input and outputs the features $f(t_i)$ of each 3D point. Similar to color, volume rendering can be used to render the 3D feature field into the 2D image as:
\begin{equation}
\label{eq:feature_render}
    \hat{F}(t) =\sum_{i=1}^n T(t_i)\alpha(t_i)f(t_i)
\end{equation}
To optimize the 3D SAM feature field $f$, we adopt the L1 loss between the rendered features $\hat{F}(t)$ and SAM encoder features $Enc_{I}(I)$:
\begin{equation}
\label{eq:feature_loss}
    L_{f}=\frac{1}{R}\sum\limits_{r}\left\|\hat{F}(r)-Enc_{I}(I)\right\|_1
\end{equation}
As for the color branch and SDF branch, we follow the previous work and adopt photometric loss and Eikonal loss~\cite{gropp2020implicit} to supervise their training respectively, which can be defined as:
\begin{equation}
\label{eq:render_loss}
\begin{split}
    L_{c}&=\frac{1}{R}\sum\limits_{r}\left\|\hat{C}(r)-C(r)\right\|_2^2 \\
    L_{eik}&=\frac{1}{R \cdot N_s}\sum\limits_{r,i}(|n|-1)^2
\end{split}
\end{equation}
Finally, we minimize the weighted sum of the above losses:

\begin{equation}
\label{eq:total_loss}
L_{total}=L_{c}+\lambda_{eik} L_{eik}+\lambda_{f} L_{f}+\lambda_{v} L_{v}
\end{equation}

\begin{table*}[htb]
\caption{Quantitative comparisons on masks generated by our iterative training. $M_o$ represents masks generated by 3D occupancy field. $M_{SAM}$ represents masks generated by SAM with prompts from 3D occupancy field. For baselines, $M_{Proj}$, $M_{ISRF}$, $M_{DINO}$ and $M_{SA3D}$ indicate masks from simply prompts projection, ISRF, DINO used by SPIn-NeRF and SA3D. Mean represents the average IoU.}
\centering
\resizebox{0.85\textwidth}{!}{
    \begin{tabular}{l|ccccccccccccccc|c}
    \toprule
    \textbf{Scan ID} &24 &37 &40 &55 &63 &65 &69 &83 &97 &105 &106 &110 &114 &118 &122 &Mean\\
    $M_{Proj}$       & 0.541 & 0.598 & 0.512 & 0.495 & 0.555 & 0.582 & 0.509 & 0.317 & 0.407 & 0.687 & 0.631 & 0.547 & 0.521 & 0.498 & 0.432 & 0.522 \\
    $M_{ISRF}$       & 0.846 & 0.765 & 0.715 & 0.558 & 0.717 & 0.787 & 0.687 & 0.692 & 0.231 & 0.578 & 0.761 & 0.687 & 0.641 & 0.690 & 0.549 & 0.660 \\
    $M_{DINO}$       & 0.941 & 0.961 & 0.952 & 0.972 & 0.945 & 0.959 & 0.943	& 0.897 & 0.934 & 0.932 & 0.937 & 0.976 & 0.937 & 0.957 & 0.925 & 0.945 \\
    $M_{SA3D}$       & 0.852 & 0.946 & 0.934 & 0.954 & 0.921 & 0.897 & 0.936	& 0.841 & 0.947 & 0.942 & 0.940 & 0.909 & 0.919 & 0.961 & 0.946 & 0.923 \\
    \midrule
    $M_o$       & 0.935 & 0.951 & 0.963 & 0.975 & 0.982 & 0.983 & 0.963 & 0.974 & 0.979 & 0.962 & 0.955 & 0.943 & 0.925 & 0.937 & 0.967 & \textbf{0.960} \\
    $M_{SAM}$   & 0.962 & 0.986 & 0.991 & 0.991 & 0.991 & 0.986 & 0.971 & 0.984 & 0.987 & 0.992 & 0.965 & 0.990 & 0.990 & 0.968 & 0.984 & \textbf{0.983} \\
    \bottomrule
    \end{tabular}
}
\label{tab:mask_qal}
\end{table*}

\section{Experiments}

We conduct extensive experiments to evaluate the effectiveness of the proposed method for neural 3D target object reconstruction. We first describe the experimental settings and then compare our method with the SOTA approaches on the DTU dataset~\cite{DTU2014}. We also provide the qualitative analysis on the LLFF dataset~\cite{mildenhall2019llff}. We further conduct a comprehensive ablation study to evaluate the contribution of each component. Due to the space limitation, we provide more reconstruction results on the BlendedMVS dataset~\cite{yao2020blendedmvs} in the supplementary material.

\subsection{Experimental Settings}
\noindent\textbf{Datasets.} 
For \textbf{Quantitative comparison}, following the previous works~\cite{wang2021neus}, we evaluate our proposed method on the selected 15 scenes from the DTU dataset. Specifically, there are 64 or 49 images with $1600\times 1200$ image resolution in each scene. Since the DTU dataset provides ground truth foreground masks, we first compare the segmentation generated by NTO3D and baselines. Then we evaluate the rendering and reconstruction quality from two aspects: training with mask supervision (w/ mask) and without mask supervision (w/o mask).
For \textbf{Qualitative comparison}, following the previous work~\cite{mildenhall2020nerf}, we show visualizations on 9 challenging scenes from the LLFF dataset. There are 20 to 62 images with a fixed image resolution of $1008\times 756$ in each scene, and we randomly select 1/8 of the entire images to construct the test set. Since LLFF does not contain ground truth foreground masks, we first manually annotate a target object in the scene and compare the visual quality.

\noindent\textbf{Implementation Details.}
We follow the implementation details specified in NeuS~\cite{wang2021neus} and Instant-NSR~\cite{zhao2022human}. we adopt the network architecture of Instant-NSR, which consists of two MLPs and the multi-resolution hash table to encode SDF and color, respectively. We utilize the Adam optimizer~\cite{kingma2014adam} with $(\beta_1, \beta_2) = (0.9, 0.999)$ to update our neural networks, and the learning rates warm up from 0 to $1\times 10^{-3}$ in the first 5k iterations and then controlled by the linear decay scheme to the latest learning rate of $1\times 10^{-5}$. We set the number of rays to 4096 and sample 80 points for each ray. We first train the neural field of the scene and then learn the 3D occupancy field with 1k interval iterations. Finally, we finetune NTO3D for 20k iterations along with the 3D feature field. All experiments are conducted on NVIDIA 3090 GPUs. More implementation details are reported in the supplementary material.

\begin{table*}[htb]
\caption{Quantitative comparisons with other methods on the task of novel view synthesis. Mean represents the average value of PSNR and SSIM.}
\centering
\setlength\tabcolsep{2pt}
\resizebox{0.80\linewidth}{!}{
\begin{tabular}[t]{l|ccccccccccccccc|c}
            \toprule
            \textbf{Scan ID} &
            24 & 37 & 40 & 55 & 63 & 65 & 69 & 83 & 97 & 105 & 106 & 110 & 114 & 118 & 122 & Mean\\
            \midrule
            \multicolumn{17}{c}{\color{gray}{\textit{Train w/o mask setting}} }       \\
            PSNR\small{(NeuS)} & 23.46 & 26.01 & 28.34 & 22.41 & 14.99 & 16.66 & 24.09 & 15.37 & 19.87 & 15.10 & 26.74 & 24.73 & 28.91 & 35.92 & 35.20 & 23.85
            \\
            PSNR\small{(NeRF)} & 21.13 & 23.00 & 18.91 & 22.97 & 25.55 & 12.7 & 22.76 & 25.19 & 28.54 & 35.29 & 18.31 & 17.91 & 21.44 & 21.48 & 22.68 & 22.52 \\
            \midrule
            SSIM\small{(NeuS)} & 0.876 & 0.611 & 0.911 & 0.903 & 0.897 & 0.744 & 0.844 & 0.840 & 0.856 & 0.826 & 0.838 & 0.871 & 0.914 & 0.968 & 0.972 & 0.858 \\
            SSIM\small{(NeRF)} & 0.791 & 0.860 & 0.817 & 0.853 & 0.921 & 0.798 & 0.829 & 0.938 & 0.932 & 0.960 & 0.757 & 0.742 & 0.826 & 0.781 & 0.821 & 0.842 \\
            \midrule
            \multicolumn{17}{c}{\color{gray}{\textit{Train w/ mask setting}} }       \\
            PSNR\small{(NeuS)} & 28.38 & 23.75 & 30.47 & 29.52 & 29.83 & 32.29 & 29.01 & 32.28 & 28.39 & 29.43 & 32.26 & 36.15 & 30.58 & 36.22 & 33.60 & 30.81 \\
            PSNR\small{(Ours)} & 28.71 & 26.96 & 30.92 & 31.87 & 34.11 & 32.92 & 31.61 & 36.98 & 28.07 & 33.10 & 35.98 & 36.85 & 32.64 & 37.69 & 37.55 & \textbf{33.06} \\
            \midrule
            SSIM\small{(NeuS)} & 0.887 & 0.891 & 0.917 & 0.958 & 0.951 & 0.969 & 0.935 & 0.969 & 0.920 & 0.924 & 0.954 & 0.970 & 0.942 & 0.969 & 0.968 & 0.942 \\
            SSIM\small{(Ours)} & 0.897 & 0.913 & 0.925 & 0.968 & 0.958 & 0.956 & 0.948 & 0.969 & 0.920 & 0.956 & 0.979 & 0.973 & 0.967 & 0.969 & 0.980 & \textbf{0.952} \\
            \bottomrule
            \end{tabular}
}
\label{tab:PSNR}
\end{table*}

\begin{table*}[htb]
\caption{Chamfer distances comparisons with other methods on the DTU dataset. COLMAP results are achieved by trim=0.}
\centering
\setlength\tabcolsep{2pt}
\resizebox{0.70\linewidth}{!}{
\begin{tabular}[t]{l|ccccccccccccccc|c}
\toprule
            \textbf{Scan ID} &
            24 & 37 & 40 & 55 & 63 & 65 & 69 & 83 & 97 & 105 & 106 & 110 & 114 & 118 & 122 & Mean\\
            \midrule
            \multicolumn{17}{c}{\color{gray}{\textit{Train w/o mask setting}} }       \\
            Post-Segmentation & 0.98 & 1.20 & 1.58 & 2.07 & 2.09 & 2.94 & 1.02 & 2.56 & 2.64 & 1.14 & 1.42 & 1.21 & 1.01 & 1.36 & 1.24 & 1.63 \\
            COLMAP & 0.81 & 2.05 & 0.73 & 1.22 & 1.79 & 1.58 & 1.02 & 3.05 & 1.40 & 2.05 & 1.00 & 1.32 & 0.49 & 0.78 & 1.17 & 1.36 \\
            NeRF & 1.90 & 1.60 & 1.85 & 0.58 & 2.28 & 1.27 & 1.47 & 1.67 & 2.05 & 1.07 & 0.88 & 2.53 & 1.06 & 1.15 & 0.96 & 1.49\\
            UNISURF & 1.32 & 1.36 & 1.72 & 0.44 & 1.35 & 0.79 & 0.80 & 1.49 & 1.37 & 0.89 & 0.59 & 1.47 & 0.46 & 0.59 & 0.62 & 1.02\\
            NeuS & 1.00 & 1.37 & 0.93 & 0.43 & 1.10 & 0.65 & \textbf{0.57} & 1.48 & 1.09 & 0.83 & 0.52 & 1.20 & \textbf{0.35} & 0.49 & 0.54 & 0.84\\
            \midrule
            \multicolumn{17}{c}{\color{gray}{\textit{Train w/ mask setting}} }       \\
            Prompts Projection & 1.21 & 1.14 & 1.47 & 2.37 & 1.60 & 2.01 & 1.27 & 1.03 & 3.28 & 1.54 & 1.47 & 1.20 & 1.03 & 1.54 & 1.21 & 1.56 \\
            ISRF & 1.02 & 1.34 & 0.86 & 1.45 & 1.36 & 1.00 & 1.12 & 1.58 & 2.07 & 1.20 & 0.98 & 1.36 & 1.21 & 0.98 & 0.87 & 1.23 \\
            SPIn-NeRF & 0.97 & 1.12 & 0.76 & 0.96 & 1.77 & 1.02 & 1.25 & 1.50 & 2.05 & 0.98 & 0.67 & 1.25 & 0.96 & 0.59 & 0.51 & 1.02 \\
            SA3D & 0.95 & 1.09 & 0.82 & 1.03 & 1.63 & 1.23 & 1.14 & 1.41 & 2.13 & 0.95 & 0.73 & 1.29 & 0.99 & 0.72 & 0.56 & 1.11 \\
            IDR & 1.63 & 1.87 & 0.63 & 0.48 & 1.04 & 0.79 & 0.77 & 1.33 & 1.16 & 0.76 & 0.67 & \textbf{0.90} & 0.42 & 0.51 & 0.53 & 0.90 \\
            NeuS & 0.83 & \textbf{0.98} & \textbf{0.56} & 0.37 & 1.13 & 0.59 & 0.60 & 1.45 & 0.95 & 0.78 & 0.52 & 1.43 & 0.36 & 0.45 & \textbf{0.45} & 0.77 \\
            Ours &\textbf{0.82} & 1.14 & 0.60 & \textbf{0.35} & \textbf{1.01} & \textbf{0.53} & 0.63 & \textbf{1.31} & \textbf{0.86} & \textbf{0.73} & \textbf{0.51} & 1.15 & 0.45 & \textbf{0.42} & 0.46 & \textbf{0.73}\\
            \bottomrule
            \end{tabular}
}
\label{tab:CD}
\end{table*}

\subsection{Quantitative Comparison}
In this section, we first report the comparison of generated segmentation masks. We compare against the segmentation masks generated by SPIn-NeRF~\cite{spinnerf}, ISRF~\cite{isrfgoel2023} and SA3D~\cite{cen2023segment}. We also compare a simple baseline named Prompts Projection: We directly project prompts from a view to other views and use SAM to generate the 2D masks based on the projected prompts. Then we report the comparisons of neural rendering and reconstruction.
We compare our method with IDR~\cite{yariv2020multiview}, NeRF~\cite{mildenhall2020nerf}, COLMAP~\cite{schonberger2016pixelwise}, UNISURF~\cite{oechsle2021unisurf}, and NeuS~\cite{wang2021neus}. We also compare a simple baseline named Post-processing Segmentation: We first reconstruct the whole scene and then use ground truth 2D masks provided by DTU to segment the desired object. 
It should be noted that NTO3D is designed for neural reconstruction similar to NeuS based on the SDF field, while ISRF, SPIn-NeRF and SA3D~\cite{cen2023segment} are all designed for neural rendering based on the density field. To make a fair comparison, we first project the 2D foreground masks of ISRF, SPIn-NeRF and SA3D to 3D space and segment the 3D target object based on the reconstruction of NeuS.

\begin{figure*}[ht!]
    \centering
    \includegraphics[width=0.80\textwidth]{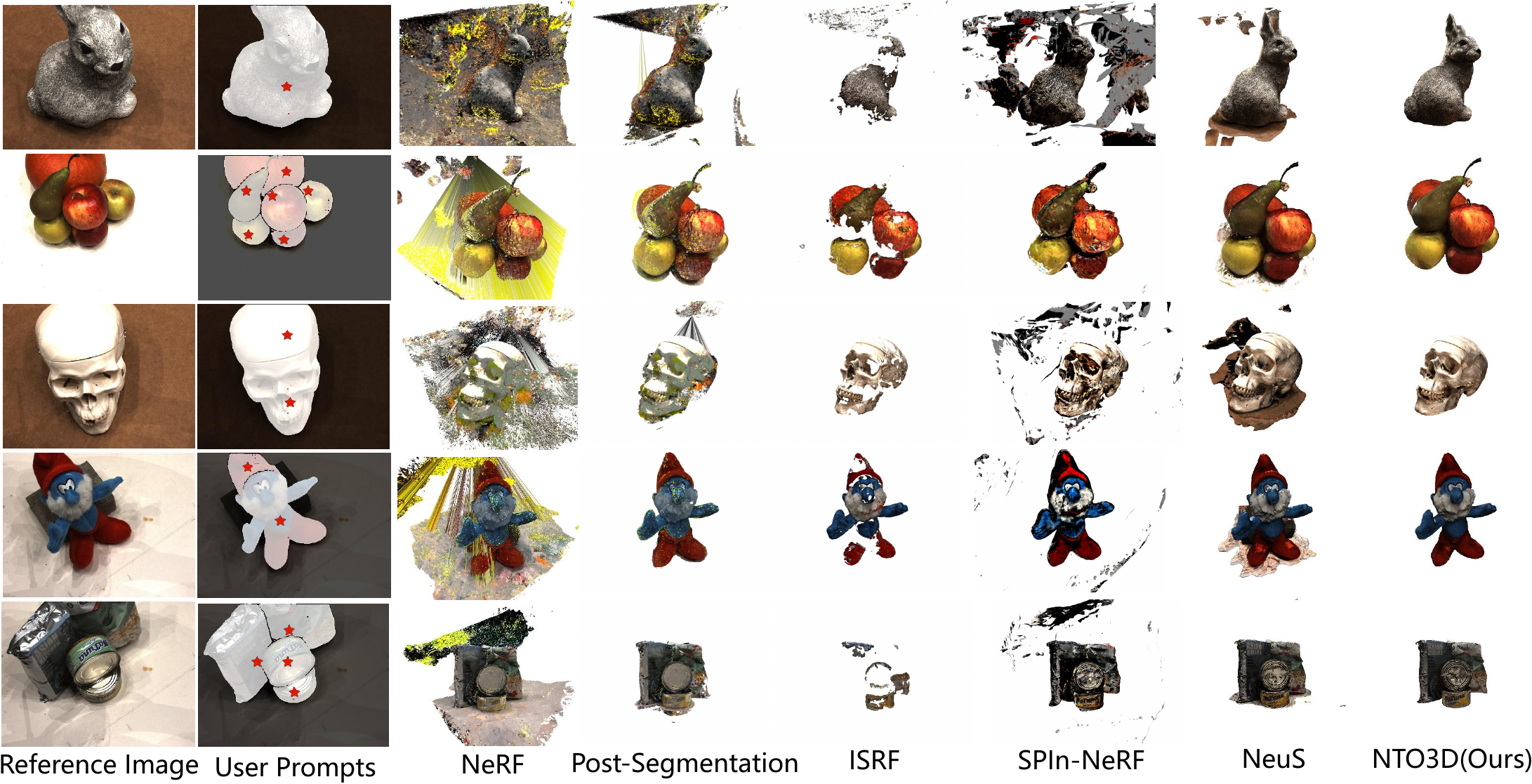}
    \caption{Qualitative comparison on DTU. Best viewed in colors.}
    \label{fig:DTU}
\end{figure*}

\begin{figure*}[ht!]
    \centering
    \includegraphics[width=0.80\textwidth]{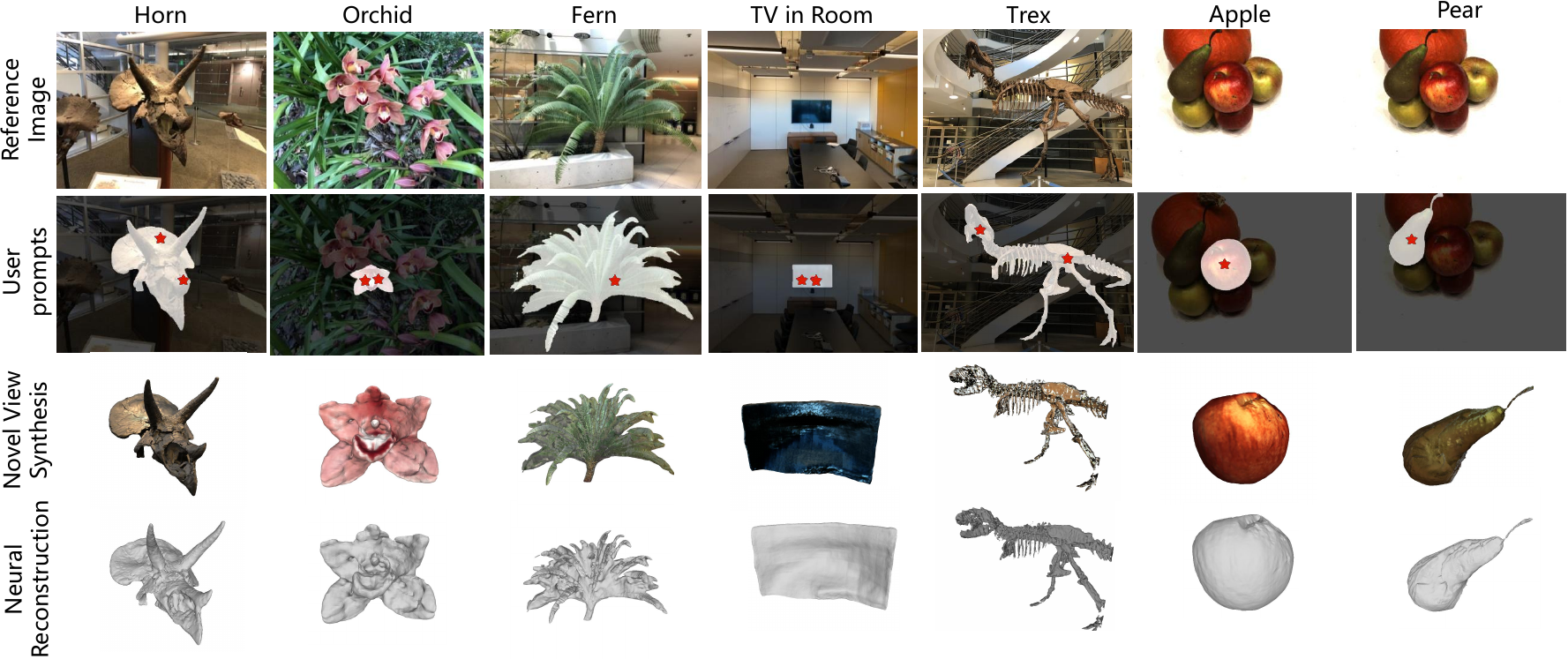}
    \caption{Qualitative comparison on LLFF and fruit scene in DTU. These scenes have the following characteristics: the foreground objects consist of multiple independent objects, and the background is more complex. Best view in colors.}
    \label{fig:LLFF_visual}
\end{figure*}

\noindent\textbf{Segmentation Comparison.} As shown in Tab.~\ref{tab:mask_qal}, after iterative training, NTO3D achieves high segmentation mask quality. On one hand, the precise mask generated by SAM guides the 3D occupancy field in distinguishing foreground and background voxels effectively. On the other hand, the 3D occupancy field aggregates the multi-view 2D segmentation masks, enabling the generation of cross-view prompts even with only a single view annotated by users.

\noindent\textbf{Rendering Comparison.} We compare NTO3D with the previous SOTA volume rendering approaches. We held out 10\% of the images in the DTU dataset as the testing set and the others as the training set. During training, we split the baselines into two settings: train w/o ground truth masks and w/ ground truth masks. For NTO3D, we use masks generated by SAM after iteration lifting for training. We compare the rendering quality on the testing set with masks regarding PSNR and SSIM. 
During the test, we calculate the metrics between prediction and masked ground truth. As shown in Tab.~\ref{tab:PSNR}, baselines trained without object masks render lower quality than those trained with object masks. Our method shows significant improvement in PSNR and SSIM, demonstrating that our method can aggregate the multi-view 2D segmentation masks to improve the novel view synthesis quality of a target object.

\noindent\textbf{Reconstruction Comparison.} We also measure the reconstruction quality with the Chamfer distances and compare NTO3D with other methods, as shown in Tab.~\ref{tab:CD}. 
Similar to the rendering comparison, training without masks introduces more background noises into the neural field. 
Simply projecting point prompts to other views leads to bad performance. This is because prompts projection heavily relies on the quality of depth, while the reconstructed depth of NeuS is not accurate enough. Therefore, SAM cannot correctly segment the target object. With the help of the 3D occupancy field and the 3D feature field, our approach reduces the Chamfer distance to 0.73 and outperforms the baseline methods. The results demonstrate that neural reconstruction can benefit from the NTO3D pipeline.

\subsection{Qualitative Analysis}

We conduct the qualitative comparisons on DTU and LLFF datasets. As shown in Fig.~\ref{fig:DTU} and Fig.~\ref{fig:LLFF_visual}, we provide reference images and prompts indicated by users on the initial view. Due to the space limitation, please refer to supplementary materials for qualitative results on BlendedMVS, which contains more complex and larger scenes.

On DTU datasets, we compare the visualization quality of NTO3D with other methods. NeRF shows the worst visualization quality since it reconstructs the whole scene. Post-Segmentation also leads to bad performance since reprojecting 2D segmentation masks provided by DTU back into 3D does not completely remove the background. ISRF offers a tool for interactive segmentation, but masks extracted by ISRF are sensitive to a bunch of parameters. Inappropriate parameters result in lower-quality masks for ISRF. SPIn-NeRF uses DINO for segmentation, which treats multi-view inputs as a video sequence. However, the segmentation accuracy of SPIn-NeRF is worse than NTO3D. SA3D uses an inverse rendering for mask generation but still exhibits lower segmentation accuracy compared to NTO3D, which results in worse chamfer distance.
Although NeuS contains a background model, which helps it focus on the foreground objects, its reconstruction results still inevitably show the background near the target objects.
Thanks to the proposed 3D occupancy field, NTO3D can generate high-quality masks of foreground objects without tedious annotation on all views. Additionally, we can witness that the NTO3D reconstructs higher surface quality with the help of 3D SAM feature fields.

On LLFF datasets, we choose one object for each scene as the target object. We can see that whether the selected object is significant or not in the scene, NTO3D can segment the target object based on the user's prompts and obtain impressive reconstruction quality. Besides, we also provide the results of one object among several foreground objects in the last two columns. This further demonstrates that with the help of NTO3D, we are able to reconstruct any target objects of the scene.

\vspace{-0.2cm}
\begin{table}[ht!]
\caption{\textbf{Ablation Study of NTO3D}. CD indicates the Chamfer distance.}
\vspace{-0.3cm}
\small
\centering
\resizebox{0.80\linewidth}{!}{
	\begin{tabular}{lcll}
	\toprule
		Variant &PSNR$\uparrow$ &SSIM$\uparrow$ & CD$\downarrow$\\
		\cmidrule(lr){1-1} \cmidrule(lr){2-4}
      Instant-NSR &29.94 &0.8914 &0.82\\
      \cmidrule(lr){1-4}
      + 3D Occupancy Field &32.44 &0.9319 &0.76\\
      + 3D SAM Feature Field &30.76 &0.9136 &0.78\\
      \cmidrule(lr){1-4}
      NTO3D (our) &33.06 &0.9520 &0.73 \\
	  \bottomrule
	\end{tabular}
}
\label{t3}
\label{tab:ablation}
\end{table}
\vspace{-0.4cm}
\subsection{Ablation study}

\noindent\textbf{Evaluation of Each Component.} We study the effectiveness of the proposed 3D occupancy field and 3D feature field. The experiments are done on the DTU dataset and average the results of all scenes. As shown in Tab.~\ref{tab:ablation}, with the aid of the 3D occupancy field, our method avoids the influence of the background and focuses on the reconstruction of target objects. Since the SAM encoder contains abundant knowledge, the 3D feature field helps to boost the reconstruction quality. With the proposed two contributions, NTO3D can efficiently segment and reconstruct the target object indicated by users in the scene.

\noindent\textbf{Overlap Ratio Between Views.} We further explore the effects of the max overlap ratio between views to study the choice of the initial view. We conduct the experiments on a random scene of the DTU dataset and calculate the Max View Distance and the Max Overlap Ratio between the first manually annotated view and other views. As shown in Tab.~\ref{tab:Influence of overlap ratio}, the final reconstruction accuracy is approximately the same for different initial views with different overlap ratios. Since the proposed 3D occupancy field iteratively obtains foreground-background segmentation in 3D space, NTO3D can avoid confusion from different views. Once the initial view appropriately segments the target object, NTO3D can always successfully reconstruct target objects.

\begin{table}[ht!]
\caption{Influence of overlap ratio on reconstruction quality.}
\vspace{-0.2cm}
\small
\centering
\resizebox{0.90\linewidth}{!}{
	\begin{tabular}{c|ccccc}
	\toprule
        Initial View & 1 & 5\% & 10\% & 20\% & 30\% \\
		\cmidrule(lr){1-6}
        Max View Distance &103.63° &122.87° &108.5° &110.09° &101.49° \\
        \cmidrule(lr){1-6}
        Max Overlap Ratio &0.471 &0.149 &0.500 &0.250 &0.322 \\
        \cmidrule(lr){1-6}
        Chamfer Distance$\downarrow$ &0.731 &0.742 &0.735 &0.740 &0.738 \\
	  \bottomrule
	\end{tabular}
}
\vspace{-0.4cm}
\label{tab:Influence of overlap ratio}
\end{table}

\noindent\textbf{Different Stages of Distilling SAM Features.} We conduct the experiments of distilling SAM features at different stages: 1). pretraining before the first step of our method (denoted by pretrain). 2). iterative training for 3D occupancy field (denoted by iterative). 3). finetuning on the target object (our default setting, denoted by finetune). We conduct the experiments on a random scene of the DTU dataset and report the mIoU and chamfer distance. As shown in Tab.~\ref{tab:Different stages}, distilling SAM features at pretraining and iterative training performs slightly worse than distilling at finetuning. Without the foreground mask, SAM features contain the background information, which might slightly degrade the performance of target object reconstruction. In addition, encoding images with SAM requires additional computational cost, therefore it is unnecessary to distill SAM features at early stages.
\vspace{-0.2cm}
\begin{table}[ht!]
\caption{Different stages of distilling SAM features.}
\vspace{-0.2cm}
\small
\centering
\resizebox{0.80\linewidth}{!}{
	\begin{tabular}{c|ccc}
	\toprule
        Experiment Setting & pretrain & iterative & finetune \\
		\cmidrule(lr){1-4}
        mIoU &0.981 &0.980 &0.991 \\
        \cmidrule(lr){1-4}
        Chamfer Distance &0.360 &0.358 &0.353 \\
	  \bottomrule
	\end{tabular}
}
\label{tab:Different stages}
\end{table}
\vspace{-0.6cm}
\section{Limitation}
\vspace{-0.2cm}
Although SAM has a powerful segmentation ability, when facing challenging scenes, it is also difficult to segment the target object~\cite{ji2023segment}. If SAM fails to segment target objects with available prompts, our method will fail to learn the 3D occupancy field. A possible solution is to fine-tune SAM on challenging scenes with parameter-efficient-finetuning (PEFT) techniques. Despite the limitations, NTO3D has demonstrated the potential of combining large foundation models and neural fields.
\vspace{-0.4cm}
\section{Conclusion}
\vspace{-0.2cm}
To reconstruct a certain object indicated by users on-the-fly and boost the reconstruction quality, the paper applies the Segment Anything Model to help 3D object reconstruction. The proposed method Neural Target Object 3D Reconstruction (NTO3D) first leverages a 3D occupancy field to lift the multi-view 2D segmentation masks generated by SAM. With the help of 3D occupancy field, NTO3D is able to segment target objects and eliminate background interference. To boost the reconstruction quality, we further propose a 3D SAM Feather Field to lift pixel-level features into voxel space. Finally, we conduct several experiments on several datasets and demonstrate NTO3D can obtain better reconstruction quality. Please refer to supplementary materials for more results.

\textbf{Acknowledgement.} Shanghang Zhang is supported by the National Science and Technology Major Project of China (No. 2022ZD0117801).

{
    \small
    \bibliographystyle{ieeenat_fullname}
    \bibliography{main}
}


\end{document}

%% file: preamble.tex
\usepackage[dvipsnames]{xcolor}
